\documentclass[10pt, a4paper]{article}
\usepackage{lrec2022} 
\usepackage{multibib}
\newcites{languageresource}{Language Resources}
\usepackage{graphicx}
\usepackage{tabularx}
\usepackage{soul}
\usepackage{booktabs}
\usepackage{titlesec}

\titleformat{\section}{\normalfont\large\bfseries\center}{\thesection.}{1em}{}
\titleformat{\subsection}{\normalfont\SmallTitleFont\bfseries\raggedright}{\thesubsection.}{1em}{}
\titleformat{\subsubsection}{\normalfont\normalsize\bfseries\raggedright}{\thesubsubsection.}{1em}{}
\renewcommand\thesection{\arabic{section}}
\renewcommand\thesubsection{\thesection.\arabic{subsection}}
\renewcommand\thesubsubsection{\thesubsection.\arabic{subsubsection}}

\usepackage{epstopdf}
\usepackage[utf8]{inputenc}

\usepackage{hyperref}
\usepackage{xstring}

\usepackage{color}

\usepackage{todonotes}
\usepackage{caption}
\usepackage{subcaption}

\title{PLOD: An Abbreviation Detection Dataset for Scientific Documents}

\name{Leonardo Zilio\textsuperscript{1}, Hadeel Saadany\textsuperscript{1}, Prashant Sharma\textsuperscript{2}, \\
{\bf \large Diptesh Kanojia\textsuperscript{1,3}, Constantin Orăsan\textsuperscript{1,3}}}

\address{\textsuperscript{1}Centre for Translation Studies, University of Surrey, United Kingdom. \\
\textsuperscript{2}Hitachi CRL, Japan.\\
\textsuperscript{3}Surrey Institute for People-centred AI, University of Surrey, United Kingdom. \\
\textsuperscript{1,3}\{l.zilio, h.saadany, d.kanojia, c.orasan\}@surrey.ac.uk, 
\textsuperscript{2}prashaantsharmaa@gmail.com\\
}

\abstract{
The detection and extraction of abbreviations from unstructured texts can help to improve the performance of Natural Language Processing tasks, such as machine translation and information retrieval. However, in terms of publicly available datasets, there is not enough data for training deep-neural-networks-based models to the point of generalising well over data. This paper presents PLOD, a large-scale dataset for abbreviation detection and extraction that contains 160k+ segments automatically annotated with abbreviations and their long forms. We performed manual validation over a set of instances and a complete automatic validation for this dataset. We then used it to generate several baseline models for detecting abbreviations and long forms. The best models achieved an F1-score of 0.92 for abbreviations and 0.89 for detecting their corresponding long forms. We release this dataset along with our code and all the models publicly in \href{https://github.com/surrey-nlp/PLOD-AbbreviationDetection}{this Github Repository}. \\ \newline }

\begin{document}

\maketitleabstract

\section{Introduction}
\label{sec:intro}

A pervasive characteristic of scientific reports and research papers is their frequent use of abbreviations~\cite{wu2011detecting}. For submitting to various journals, authors are also required to submit an abbreviation glossary, \textit{i.e.}, a list of short forms with their expanded long forms. Such a glossary is essential for the reader to understand the domain-specific terminology used in the reported work. 

From a linguistic point of view, there are different typologies of abbreviations, and often authors disagree in relation to a common classification system~\cite{fabijanic2015dictionary}. As~\newcite{tchiotashvili2021classification} explain, there are authors advocating for the distinction between initialisms and acronyms\footnote{According to those authors, initialisms and acronyms are both short forms that are created using the initial letter(s) from a sequence of tokens, and are differentiated according to how they are pronounced.}, and others that defend a separation between shortening abbreviations and initial abbreviations. Considering that a typology of abbreviations is not our focus in this paper, we will use the terms ``short form'', ``abbreviation'' and ``abbreviated token'' interchangeably and as umbrella terms to denote any token(s) that can be expanded into a longer token or into a sequence of tokens that corresponds to its long form. As such, unless explicitly stated, we will not differentiate between, for instance, abbreviations and acronyms, and nor does our dataset.

From a Natural Language Processing (NLP) point of view, abbreviations are problematic for automatic processing, and the presence of short forms might hinder the machine processing of unstructured text. For example, a machine translation system may not provide a suitable translation for such tokens. Abbreviated tokens can pose a problem for almost any NLP system, because they often contain important information, such as names of diseases, drug names, or common procedures which must be recognisable in the translated document. The performance of information retrieval can also be affected in terms of both precision and recall due to incorrect abbreviation expansion~\cite{toole2000hybrid}. Therefore, the detection and extraction of accurate abbreviated tokens and their corresponding long forms is an important task that can significantly impact NLP systems' output. 

Any NLP system which attempts to extract such information from unstructured text faces several challenges, because abbreviations:

\begin{itemize}
    \item are domain-specific (\textit{e.g.}, BMI can mean ``Body Mass Index'' or ``Bilinear Matrix Inequalities'', depending on the domain or the context);
    \item often have ambiguous connotations (\textit{e.g.}, CI can mean ``conditional independence'', ``confidence interval'', or ``compound interest''; all from the same domain); 
    \item can often contain sub-abbreviations which are not fully expanded in the immediate context (\textit{e.g.}, NMT might only contain the expansion ``Neural MT'', where MT can be found expanded earlier in the document);
    \item can have multiple letters that are a part of the same word (\textit{e.g.}, subsequence kernel (SSK) or maximum entropy (MaxEnt)); and
    \item can often appear in a text unaccompanied by their respective long forms. 
\end{itemize}

The challenges discussed here also show that rule-based approaches will fail to perform well at this task as they will try to generalise over a pattern or a regular expression to detect abbreviations from a text. Multiple outliers cannot be detected with the help of such approaches. Therefore, it is important to create robust NLP systems that can detect and extract abbreviations with their corresponding long forms. The detection of short and long forms can help automate glossary generation for researchers. Similarly, it can help the expansion of the abbreviations in a free text, thus enabling downstream NLP tasks like Machine Translation (MT) or Information Retrieval (IR) to perform better.

In this paper, we describe our efforts to collect a large dataset of abbreviations from online open-source journals and create baseline NLP systems for the task of abbreviation detection. We present our new PLOD dataset for abbreviation detection and provide a detailed analysis of its main features. We also describe steps for dataset crawling and cleaning, and for its manual and automatic validation. With the help of various publicly available language models, we perform fine-tuning to create baseline models for abbreviation detection based on the PLOD dataset. 
We also test our baseline models performance on another publicly available acronym-extraction dataset to show their efficacy on this sister task. Our contributions with this paper are summarised below:
\begin{itemize}
    \item We present PLOD, a large dataset for the detection and extraction of short and long forms.
    \item We provide several pre-trained baseline models that are readily available to use.
\end{itemize}

The rest of this paper is organised as follows: in Section~\ref{sec:relwork} we discuss existing efforts for the extraction of abbreviations; Section~\ref{sec:resource} describes the methodology applied for the creation and validation of the new dataset and also presents some of its main statistics at the end; Section~\ref{sec:resourceeval} contains an extrinsic evaluation of the resource, where we created several baseline models to test the dataset for automatically detecting short and long forms; in Section \ref{sec:results} we detail the results of each baseline model; finally, Section \ref{sec:conc} briefly summarises what was achieved with this research and presents an overview of future steps.




\section{Related Work} 
\label{sec:relwork}

For many years, researchers have employed machine-learning-based methods to detect abbreviations from generic English texts. \newcite{toole2000hybrid} introduces a hybrid two-stage approach for the identification and expansion of abbreviations based on a dataset from the Air Safety Reporting System (ASRS) database. The author proposes various features and utilises a binary decision tree to model the characteristics of an abbreviation. Similarly,~\newcite{vanopstal2010towards} use Support Vector Machine (SVM) to classify abbreviations based on various features and on a dataset in the medical domain. Abbreviation detection has been more popular in the clinical and the medical domains as a lot of unstructured free text is prevalent in these areas which also contains multiple abbreviations.~\newcite{xu2009methods} also propose a decision-tree-based approach for the classification of abbreviations from clinical narratives. Another research paper~\cite{wu2011detecting} shows a more exhaustive comparison of various machine-learning-based methods like decision tree, random forests, and SVM and utilise over 70 patient discharge summaries to perform the task. \newcite{kreuzthaler2016unsupervised} use an unsupervised learning approach to detect abbreviations in clinical narratives, and show a decent performance on a small German language dataset (1696 samples). More recently, the CLEF shared task for short form normalisation propelled the efforts in this area~\cite{wu2013clinical}. Another recent approach to detect abbreviations in clinical text utilises a semi-supervised learning approach to do the task~\cite{8858543} for a clinical text dataset~\cite{moon2014sense}. 

The more specific task of acronym extraction, where the focus lies on abbreviations formed by initial letter(s), has also been of interest to the NLP community and has been performed for different domains in English. Early approaches for this task were primarily rule-based~\cite{taghva1999recognizing,yeates1999automatic,park2001hybrid,larkey2000acrophile,schwartz2002simple}, but there are instances of machine learning being used for the task~\cite{nadeau2005supervised,kuo2009bioadi}. 

Recently, various deep learning-based approaches have been used for acronym extraction~\cite{rogers2021ai,li2021systems}. Similarly,~\newcite{zhu2021bert,kubal2021effective} use the fine-tuning approaches based on the recent transformer-based architectures.~\newcite{ehrmann2013acronym} show how acronym recognition patterns initially developed for medical terms can be adapted to the more general news domain. Their efforts led to automatically merging long-form variants referring to the same short form, while maintaining non-related long forms separately. Their work is based on the algorithm developed by~\newcite{schwartz2002simple}, but they perform the task of acronym extraction for 22 languages.

In fact, the acronym extraction and disambiguation shared task~\cite{veyseh-et-al-2022-Multilingual} has encouraged more participants in the area while also releasing a large-scale dataset for multilingual and multi-domain acronym extraction~\cite{veyseh-et-al-2022-MACRONYM}. However, none of the abbreviation datasets discussed can be considered significantly large for deep-learning-based approaches to generalise well enough and show decent task performance. With this work, we release a much larger dataset containing tagged abbreviations and their corresponding long forms. Although we do not make an explicit distinction, our dataset does contain, among the abbreviations, acronyms, so that the models that we present here can also help in the more specific task of acronym extraction. With the help of fine-tuning, our evaluation also shows that this dataset can help extract abbreviations and acronyms with a decent performance. 




\begin{table}[!h]
\centering
\begin{tabular}{l|c|r}
\toprule
\multicolumn{1}{c|}{\textbf{Journal}} &
  \textbf{\begin{tabular}[c]{@{}c@{}}Publication\\ Period\end{tabular}} &
  \multicolumn{1}{c}{\textbf{\begin{tabular}[c]{@{}c@{}}Number\\ of Files\end{tabular}}} \\ \toprule
PLOS Biology                                                                & 2003-present & 6,072   \\
PLOS Medicine                                                               & 2004-present & 4,494   \\
\begin{tabular}[c]{@{}l@{}}PLOS Computational\\ Biology\end{tabular}        & 2005-present & 8,473   \\
PLOS Genetics                                                               & 2005-present & 9,251   \\
PLOS Pathogens                                                              & 2005-present & 9,148   \\
PLOS Clinical Trials*                                                       & 2006-2007    & 68     \\
PLOS ONE                                                                    & 2006-present & 257,854 \\
\begin{tabular}[c]{@{}l@{}}PLOS Neglected \\ Tropical Diseases\end{tabular} & 2007-present & 9,388   \\
PLOS Currents                                                              & 2009-2018    & 697   \\ \bottomrule
\end{tabular}
*Later merged with PLOS ONE.
\caption{PLOS Journals publication period and number of files.}
\label{tab:PLOS_journals}
\end{table}

\section{Proposed Resource}
\label{sec:resource}
In this section, we discuss the new PLOD dataset that we built from  research articles published in PLOS Journals\footnote{\url{https://plos.org/}.}. We first describe the corpus that was used and the methodology for collecting data from the journals. We then describe some automatic and manual methods applied for cleaning and validating the dataset. We conclude the section with statistics of the resource we developed.

\subsection{Dataset Description and Creation}
The PLOD dataset was extracted from open access articles published in PLOS journals. The articles from these journals are freely distributed along with the PMC Open Access Subset\footnote{\url{https://www.ncbi.nlm.nih.gov/pmc/tools/openftlist/}.} and can be downloaded from their FTP server\footnote{\url{https://ftp.ncbi.nlm.nih.gov/pub/pmc/oa_bulk/}.}. The corpus contains several journals, mostly from the Biomedical domain, since 2003. All articles from these journals are written in English. Table \ref{tab:PLOS_journals} presents the areas, publication time span and number of files that we have for each PLOS journal in the corpus\footnote{The data used in this research was downloaded on 16 October 2021.}.

The full corpus contains 305,445 files (31GB) divided into several types of articles. We used only the main category, Research Articles, which accounts for 283,874 files. All articles are presented in XML format, and most of them contain a section called ``Abbreviations''. We used Python's BeautifulSoup package to process the XML files and, for each file, extracted short and long forms from the ``Abbreviations'' section and then processed all $<p>$ tags from the XML structure. The textual content inside $<p>$ tags was split into sentences using a simple regular expression, for the sake of brevity, and then each sentence was analysed to match for the occurrence of any short forms present in the paper's glossary. Where an abbreviation was found, we also looked for its long form in the same segment/textual extract. During this annotation process, we identified the indexes of the beginning and ending character in the segment. In the search for abbreviations, we used the exact same case format as it appeared in the ``Abbreviations'' section of the article, using word boundaries as delimitation (\textit{e.g.}, punctuation, brackets, spaces, apostrophes). However, for the long forms, we converted both the textual extract and the long form to lower case. We also allowed for certain plural forms in the abbreviations and in the last token of the long form (\textit{e.g.}, addition of lower cased ``s''), but these were not included in the annotation process (\textit{i.e.}, the extracted indexes and forms do not reflect these extra characters).

This resulted in a huge collection of 1,348,357 segments with annotated abbreviations. Although most papers have a glossary with abbreviations and long forms, we only found 13,883 articles in which long forms appeared together with abbreviations. This means that all of the 1.3M+ extracted segments have abbreviations, but not all of them contain long forms. When we filtered the textual extracts that had at least one long form corresponding to one of the abbreviations present in the segment, we ended up with 162,658 segments. Considering only these textual extracts that contain both short and long forms, we had a vocabulary of 56,810 unique combinations of abbreviations with their corresponding long form. This dataset was then further validated and cleaned, as we discuss in the next subsection.

\begin{table*}[!t]
\centering
\begin{tabular}{lrrr}
\toprule
\multicolumn{1}{c|}{\textbf{Journal}} &
  \multicolumn{1}{c}{\textbf{\begin{tabular}[c]{@{}c@{}}Number of\\ Segments \end{tabular}}} \vline &
  \multicolumn{1}{c}{\textbf{\begin{tabular}[c]{@{}c@{}}Annotated\\ Abbreviations\end{tabular}}} \vline &
  \multicolumn{1}{c}{\textbf{\begin{tabular}[c]{@{}c@{}}Annotated\\ Long Forms\end{tabular}}} \\ \toprule
\multicolumn{1}{l|}{PLOS Biology}                     & \multicolumn{1}{r|}{50975} & \multicolumn{1}{r|}{165099} & 97002  \\
\multicolumn{1}{l|}{PLOS Medicine}                    & \multicolumn{1}{r|}{33036} & \multicolumn{1}{r|}{83549}  & 54237  \\
\multicolumn{1}{l|}{PLOS Computational Biology}       & \multicolumn{1}{r|}{2124}  & \multicolumn{1}{r|}{4380}   & 2540   \\
\multicolumn{1}{l|}{PLOS Genetics}                    & \multicolumn{1}{r|}{2740}  & \multicolumn{1}{r|}{5659}   & 3152   \\
\multicolumn{1}{l|}{PLOS Pathogens}                   & \multicolumn{1}{r|}{2394}  & \multicolumn{1}{r|}{6225}   & 2814   \\
\multicolumn{1}{l|}{PLOS Clinical Trials}             & \multicolumn{1}{r|}{325}   & \multicolumn{1}{r|}{709}    & 410    \\
\multicolumn{1}{l|}{PLOS ONE}                         & \multicolumn{1}{r|}{69217} & \multicolumn{1}{r|}{183358} & 106031 \\
\multicolumn{1}{l|}{PLOS Neglected Tropical Diseases} & \multicolumn{1}{r|}{121}   & \multicolumn{1}{r|}{287}    & 165    \\ \hline
\textbf{Total}                                        & 160932                     & 449266                      & 266351 \\ \bottomrule
\end{tabular}
\caption{Number of annotated segments, abbreviations and long forms per journal in the PLOD dataset.}
\label{tab:annotated_segments}
\end{table*}

\subsection{Dataset Cleaning and Validation}
\label{subsec:cleaning}
Since the process of collecting abbreviations and long forms, and annotating them in textual extracts was done automatically, we did some manual checks to validate the data. In this section we describe this process of validation, elaborating on the different steps that were taken to improve the quality of our released dataset.

One of the first steps that we took was to go through 500 random examples to check overall issues with the data. During this process, we identified \textbf{two main issues}: \textbf{one-character abbreviations} were resulting in several lines where, even though both the abbreviation and long form were present in the segment, they were not connected as a textual co-reference (see Example 1); and there were \textbf{missing annotations} of abbreviations and/or long forms, either because they were not coded in the ``Abbreviations'' section of the article or because they were written differently in the text (see Example 1).

\begin{itemize}
    \item Example 1: The reaction of an \textcolor{blue}{\textbf{oligonucleotide substrate}} bearing a \textcolor{blue}{\textbf{S}} \textcolor{red}{\textit{P}}-phosphorothioate at the cleavage site (\textcolor{red}{\textit{SSp}}, Table 1) also experiences Cd2+ stimulation with the \textbf{WT} ribozyme.
\end{itemize}

In Example 1, it is possible to see that although the abbreviated token \textbf{S} and its long form \textbf{oligonucleotide substrate} are present in the segment, they are not being used as co-referents in this particular textual extract. We also see in the same example that SSp and P were not identified as abbreviations, because they were not present in the article's ``Abbreviations'' section. 

To solve the first issue, after further investigation of other similar cases, we decided to filter one-character abbreviations out of the dataset, as they were indeed a source of many issues. This resulted in the removal of 705 unique long forms from the dataset, totalling 3,877 occurrences across 1,698 segments (the filtered dataset at this point has 160,969 segments). As it will be explained in Sections \ref{sec:resourceeval} and \ref{sec:results}, we conducted our experiments using both PLOD\textsubscript{Unfiltered Dataset} and PLOD\textsubscript{Filtered Dataset} (which does not have any annotated one-character abbreviations). As for the second issue, where there is missing annotation, we decided not to act upon it, and we accepted that there will be some segments where some of the abbreviations (with or without their long forms) and/or long forms are not identified.

In a second step of validation, we used spaCy\footnote{\url{https://spacy.io/}} to create a language model specific for the annotated long forms. In this language model, any token that was not a stop-word was replaced by either a placeholder for punctuation or for content word\footnote{For the purpose of this simple language model, numbers were replaced with the same placeholder used for content words.}. This step reduced the amount of different long forms to 3,592, which allowed us to identify some oddly formed sequences, and also long forms that were too long, and possibly wrong. Based on this analysis, we annotated each segment with a number indicating whether it contains a long form that either begins or ends with stop-words or is very long (\textit{i.e.}, longer than 12 tokens), or both. This generated an extra annotation on 5,671 segments, which were not automatically excluded from the dataset, but are readily identifiable because of this annotation. In terms of the long forms that were very long, we did perform a validation among all long forms that had more than 12 tokens, and, among the 344 unique combinations of abbreviation and long form, only 17 were not correct, totalling 36 instances in the dataset. The 22 segments that contained these incorrect instances were completely removed from the final dataset. After this validation, the longest valid long form contains 26 tokens: \textit{Multicenter, Randomized, Parallel Group Efficacy and Safety Study for the Prevention of Venous Thromboembolism in Hospitalized Acutely Ill Medical Patients Comparing Rivaroxaban with Enoxaparin}; and its associated abbreviation is \textit{MAGELLAN}.

\begin{figure*}[!t]
     \centering
     \begin{subfigure}[b]{0.48\textwidth}
         \centering
         \includegraphics[width=\textwidth]{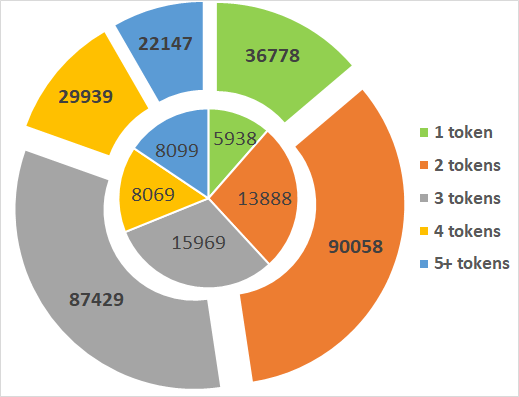}
         \caption{Length of long forms in tokens.}
         \label{fig:len_lfs}
     \end{subfigure}
     \begin{subfigure}[b]{0.48\textwidth}
         \centering
         \includegraphics[width=\textwidth]{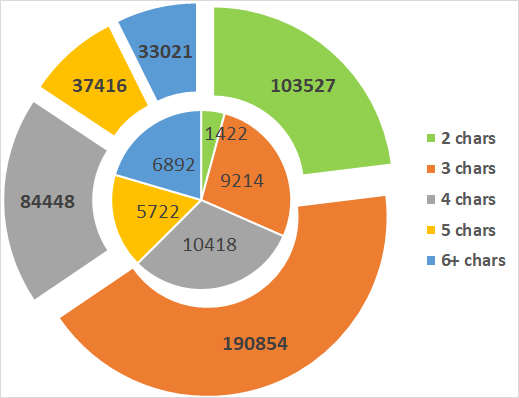}
         \caption{Length of short forms in characters (with spaces).}
         \label{fig:len_abb}
     \end{subfigure}
     \caption{Distribution of short and long forms by length in the PLOD dataset. (Internal pie charts show number of unique forms, while the external, exploded doughnut charts display the total frequency.)}
     \label{fig:length_abb_lf}
\end{figure*}

We also applied a similar validation based on the length of the abbreviations. By going through 141 instances of abbreviations that were longer than 15 characters (with spaces), we were able to identify 11 incorrect abbreviations, all of which with 19+ characters, which led to the removal of 15 segments from the final dataset. This helped us remove instances that had up to 145 characters and were clearly an error in the glossaries of the research papers from the PLOS corpus. After this validation of the long abbreviations, the longest abbreviation in the dataset has a total of 33 characters: \textit{pos regul transcr RNA pol II prom}; and it stands for \textit{positive regulation of transcription from RNA pol II promoter}.

After these two validation steps, we analysed a sample of a thousand random segments from the dataset, and there we observed that 5.5\% of the segments presented wrong annotation (\textit{i.e.}, at least one long form did not have its abbreviation as a co-reference, as we show in Example 2, where \textit{paroxon} and \textit{Pxn} are, respectively, long form and abbreviation denoting the same substance, but they are not co-referents in the segment), and 26.7\% had missing annotation (\textit{i.e.}, at least one abbreviation and/or one long form were missing from the annotated data, as we presented before in Example 1).

\begin{itemize}
    \item Example 2: Km value of \textcolor{red}{\textit{paraoxon}} towards selected \textbf{rh-PON1} mutants was nearly same while Kcat values differs however, correlates with the \textcolor{red}{\textit{Pxn}}-hydrolyzing activity of the mutants as in Fig 2.
\end{itemize}

Although there are these few issues with the dataset, it does present very useful information for abbreviation and long form detection (which is the main focus of this paper), as we will show with an extrinsic evaluation in Sections \ref{sec:resourceeval} and \ref{sec:results}, but it can also be used for typological studies related to abbreviation description. The dataset is mainly focused on biomedical texts, but it contains a representation of different types of abbreviations that can be further studied from a linguistic point of view. Besides the usual abbreviations composed by the initial letter(s) from the long form, such as \textbf{CI} (confidence interval) and \textbf{GFP} (green fluorescent protein), there are also those formed by suppressing some letters from one word, such as \textbf{TB} (tuberculosis) and \textbf{IFN} (interferon). Some short forms contain other abbreviations inside them and have a mix of upper- and lower-cased letters, such as \textbf{siRNA} (small interfering RNA) and \textbf{qPCR} (quantitative PCR), while others include parts that are not abbreviated, such as \textbf{cryo-EM} (cryo-electron microscopy). These are only a few examples of correctly annotated abbreviations that co-occur with their respective long forms in the dataset, and there are many more types and forms that can be further studied, for instance, from a Corpus Linguistics point of view. More information about quantities and types of long forms and abbreviations will be discussed in the next subsection.

\subsection{Dataset Statistics}
In this section, we describe the main statistics of the dataset that we are releasing. These statistics refer to the final PLOD\textsubscript{Filtered Dataset} after all the data was removed in the validation steps described in the previous subsection.

Table \ref{tab:annotated_segments} presents information regarding annotated segments and the amount of abbreviations and long forms divided by journal from the original PLOS corpus. It also shows the distribution of the extracted information over the different subject areas covered by the PLOS journals. As it can be seen in comparison to Table \ref{tab:PLOS_journals}, the journal \textit{PLOS Currents} is missing, as no segment was extracted from its files.

In Figure \ref{fig:len_lfs}, it is possible to have an idea of the distribution of long forms in terms of token length (split by space). Most of the long forms in the PLOD dataset have between 2 and 3 tokens, and, even though the number of unique 1- and 4-token-long long forms is representative, these forms are less frequently repeated in the texts. 

A similar distribution can be seen in Figure \ref{fig:len_abb}. It shows a higher concentration of 2- and 3-character abbreviations in terms of frequency. However, there is also an even higher number of unique 4-character abbreviations that are not as frequently used as the shorter abbreviations. 

Another interesting figure from the dataset is that the number of unique long forms is 18k+ larger than the number of unique abbreviations. This serves as an indicative of the ambiguity among the existing short forms.

After all the validation steps, the PLOD dataset was ready for an extrinsic evaluation. We then moved on to an experiment for detecting abbreviations and long forms using several pre-trained language models. The setup for this experiment is explained in the following section.

\begin{table*}[!tb]
\centering
\resizebox{\textwidth}{!}{%
\begin{tabular}{@{}lcccccccccccc@{}}
\toprule
 & \multicolumn{6}{c}{\textbf{PLOD\textsubscript{test-unfiltered}}} & \multicolumn{6}{c}{\textbf{SDU@AAAI-22 Shared Task\textsubscript{train + dev}}} \\ \midrule
 & \multicolumn{3}{c}{\textbf{Abbreviations}} & \multicolumn{3}{c}{\textbf{Long-forms}} & \multicolumn{3}{c}{\textbf{Abbreviations}} & \multicolumn{3}{c}{\textbf{Long-forms}} \\
 & P & R & F & P & R & F & P & R & F & P & R & F \\ \midrule
ALBERT\textsubscript{base} & 0.845 & 0.898 & 0.871 & 0.758 & 0.812 & \multicolumn{1}{c|}{0.784} & 0.682 & 0.638 & 0.659 & 0.462 & 0.154 & 0.231 \\
BERT\textsubscript{base-cased} & 0.855 & 0.906 & 0.880 & 0.781 & 0.826 & \multicolumn{1}{c|}{0.803} & 0.691 & 0.650 & 0.670 & 0.461 & 0.151 & 0.228 \\
DeBERTa\textsubscript{base} & 0.877 & 0.910 & 0.893 & 0.817 & 0.874 & \multicolumn{1}{c|}{0.845} & 0.682 & 0.638 & 0.659 & 0.462 & 0.154 & 0.231 \\
DistillBERT\textsubscript{base} & 0.845 & 0.900 & 0.872 & 0.772 & 0.798 & \multicolumn{1}{c|}{0.785} & 0.700 & 0.641 & 0.670 & 0.467 & 0.139 & 0.214 \\
MPNet\textsubscript{base} & 0.846 & 0.899 & 0.872 & 0.782 & 0.823 & \multicolumn{1}{c|}{0.802} & 0.691 & 0.606 & 0.645 & 0.466 & 0.145 & 0.221 \\
RoBERTa\textsubscript{base} & 0.860 & 0.919 & 0.889 & 0.805 & 0.862 & \multicolumn{1}{c|}{0.833} & \textbf{0.707} & \textbf{0.641} & \textbf{0.672} & 0.516 & 0.163 & 0.248 \\
ALBERT\textsubscript{large}  & 0.895 & 0.920 & 0.907 & 0.848 & 0.898 & \multicolumn{1}{c|}{0.872} & 0.476 & 0.607 & 0.534 & 0.397 & 0.160 & 0.228 \\
RoBERTa\textsubscript{large} & \textbf{0.911} & \textbf{0.935} & \textbf{0.922} & \textbf{0.876} & \textbf{0.921} & \multicolumn{1}{c|}{\textbf{0.898}} & 0.515 & 0.650 & 0.575 & \textbf{0.423} & \textbf{0.191} & \textbf{0.264} \\
BERT\textsubscript{large-cased} & 0.899 & 0.928 & 0.913 & 0.866 & 0.909 & \multicolumn{1}{c|}{0.887} & 0.532 & 0.645 & 0.583 & 0.362 & 0.173 & 0.234 \\ \bottomrule
\end{tabular}%
}
\caption{Results of the fine-tuning-based abbreviation detection task where \textbf{Unfiltered} data was used for training and testing. The table also shows results where we used the same trained models, but tested them on the SDU Shared Task dataset.}
\label{tab:un_filtered_results}
\end{table*}

\section{Experiment Setup for Evaluation}
\label{sec:resourceeval}

In this section, we describe the experimental procedures for generating baseline models for detecting abbreviations and long forms. This methodology also serves as an extrinsic evaluation of the PLOD dataset.

We used a customised NER pipeline from spaCy v3.2~\footnote{\href{https://spacy.io/universe/project/spacy-transformers}{spaCy Transformers}.} that utilises transformers for performing a sequence labelling task to detect abbreviations and long forms. SpaCy-transformer interoperates with PyTorch~\footnote{\href{https://pytorch.org/}{PyTorch}} and the HuggingFace transformers library~\footnote{{\href{https://huggingface.co/docs/transformers/index}{HuggingFace}}}, allowing us to access a series of pre-trained models based on state-of-the-art transformer architectures that were applied for generating our baseline models. In order to perform training with spaCy's pipeline, we annotated the PLOD dataset with an I-O-B scheme, where abbreviations were annotated as B-AB (i.e. Begin ABbreviation), and the words which were a part of the long forms were assigned B-LF (i.e. Begin Long Form) at the beginning, and I-LF (i.e. Inside Long Form) in the middle and end. This resulted in a one-token-per-line training file with the I-O-B annotation which amounted to 7,150,008 annotated tokens. We release the I-O-B-annotated dataset via a GitHub repository~\footnote{\href{https://github.com/surrey-nlp/AbbreviationDetRepo}{PLOD Dataset Github repository}.} along with the same dataset in the TSV format for researchers who wish to reproduce our experiment. We randomised and split our dataset into 70\% instances for training, 15\% for validation, and the remaining 15\% as test data. To perform comparative evaluation, we trained models on both filtered and unfiltered data (as discussed in Section~\ref{subsec:cleaning}). 

We utilised the following pre-trained Language Models (LMs) for the task of abbreviation detection: RoBERTa~\cite{liu2019roberta}, BERT~\cite{devlin-etal-2019-bert}, ALBERT~\cite{lan2019albert}, DeBERTa~\cite{he2020deberta}, DistilBERT~\cite{sanh2019distilbert}. For RoBERTa, BERT, and ALBERT, we used both \textit{base} and \textit{large} variants in our experiment. This resulted in an extensive extrinsic evaluation that was performed with the help of \textit{nine LM variants}, and with different datasets. We trained all our models with a batch size of 128 and a hidden-layer size of 64. We used a spaCy Span-Get function which transforms each batch into a list of span objects for each sentence to be processed by the transformer. This technique helps with long sentences by cutting them into smaller sequences before running the transformer and allows for overlapping of the spans to cover both left and right context. We set the span window to 128 tokens and the stride to 96 to allow for overlapping of token windows. For tokenisation we used spacy.Tokenizer.v1. For optimisation, we used Adam optimiser with an initial learning rate of 0.00001 and initial warm-up steps set to 250, with up to a total of 20000 steps. We also chose 2 Maxout units~\cite{goodfellow2013maxout} as an activation function to calculate the maximum of the inputs. These architecture parameters were chosen because they have performed well for NER tasks~\footnote{\href{https://v2.spacy.io/usage/facts-figures}{spaCy: Facts and Figures}}. The results obtained with the help of our models are presented in Tables \ref{tab:un_filtered_results} and \ref{tab:filtered_results}.

\begin{table*}[!tb]
\centering
\resizebox{\textwidth}{!}{%
\begin{tabular}{@{}lcccccccccccc@{}}
\toprule
 & \multicolumn{6}{c}{\textbf{PLOD\textsubscript{test-filtered}}} & \multicolumn{6}{c}{\textbf{SDU@AAAI-22 Shared Task\textsubscript{train + dev}}} \\ \midrule
 & \multicolumn{3}{c}{\textbf{Abbreviations}} & \multicolumn{3}{c}{\textbf{Long-forms}} & \multicolumn{3}{c}{\textbf{Abbreviations}} & \multicolumn{3}{c}{\textbf{Long-forms}} \\
 & P & R & F & P & R & F & P & R & F & P & R & F \\ \midrule
ALBERT\textsubscript{base} & 0.842 & 0.899 & 0.870 & 0.734 & 0.819 & \multicolumn{1}{c|}{0.774} & 0.716 & 0.629 & 0.670 & 0.485 & 0.146 & 0.225 \\
BERT\textsubscript{base-cased} & 0.853 & 0.902 & 0.877 & 0.766 & 0.834 & \multicolumn{1}{c|}{0.799} & 0.723 & 0.628 & 0.672 & 0.471 & 0.150 & 0.228 \\
DeBERTa\textsubscript{base} & 0.852 & 0.937 & 0.893 & 0.803 & 0.881 & \multicolumn{1}{c|}{0.840} & 0.691 & 0.606 & 0.645 & 0.466 & 0.145 & 0.221 \\
DistillBERT\textsubscript{base} & 0.842 & 0.904 & 0.872 & 0.763 & 0.805 & \multicolumn{1}{c|}{0.783} & 0.709 & 0.642 & 0.674 & 0.456 & 0.140 & 0.215 \\
MPNet\textsubscript{base} & 0.852 & 0.888 & 0.870 & 0.777 & 0.824 & \multicolumn{1}{c|}{0.800} & 0.711 & 0.586 & 0.642 & 0.472 & 0.147 & 0.224 \\
RoBERTa\textsubscript{base} & 0.857 & 0.918 & 0.886 & 0.798 & 0.867 & \multicolumn{1}{c|}{0.832} & \textbf{0.728} & \textbf{0.643} & \textbf{0.683} & \textbf{0.520} & \textbf{0.169} & \textbf{0.255} \\
ALBERT\textsubscript{large} & 0.840 & 0.918 & 0.877 & 0.770 & 0.830 & \multicolumn{1}{c|}{0.799} & 0.532 & 0.651 & 0.585 & 0.373 & 0.174 & 0.237 \\
RoBERTa\textsubscript{large}& \textbf{0.906} & \textbf{0.935} & \textbf{0.920} & \textbf{0.874} & \textbf{0.925} & \multicolumn{1}{c|}{\textbf{0.898}} & 0.502 & 0.645 & 0.564 & 0.427 & 0.181 & 0.254 \\
BERT\textsubscript{large-cased} & 0.892 & 0.931 & 0.911 & 0.858 & 0.912 & \multicolumn{1}{c|}{0.884} & 0.532 & 0.651 & 0.585 & 0.373 & 0.174 & 0.237 \\ \bottomrule
\end{tabular}%
}
\caption{Results of the fine-tuning-based abbreviation detection task where \textbf{Filtered} data was used for training and testing. The table also shows results where we used the same trained models, but tested them on the SDU Shared Task dataset.}
\label{tab:filtered_results}
\end{table*}

After training a total of 18 models both on filtered and unfiltered data, we tested them on the test splits generated from PLOD and also on the English data provided by~\newcite{veyseh-et-al-2022-Multilingual} (SDU@AAAI-22 Shared Task). Since the labelled test data from this shared task has not yet been released, we combined both the train and validation sets released by the organisers as a combined test set. With this set of experiments, we aimed to explore the efficacy of our models on a test set that is from a different domain and that contains data from other sources. Please note that the shared task data belongs to the task of Acronym Detection (AD) and our models do not make any specific distinction for acronyms, as they were trained on all sorts of short forms.




\begin{figure*}[!t]
     \centering
     \begin{subfigure}[b]{0.495\textwidth}
         \centering
         \includegraphics[width=\textwidth]{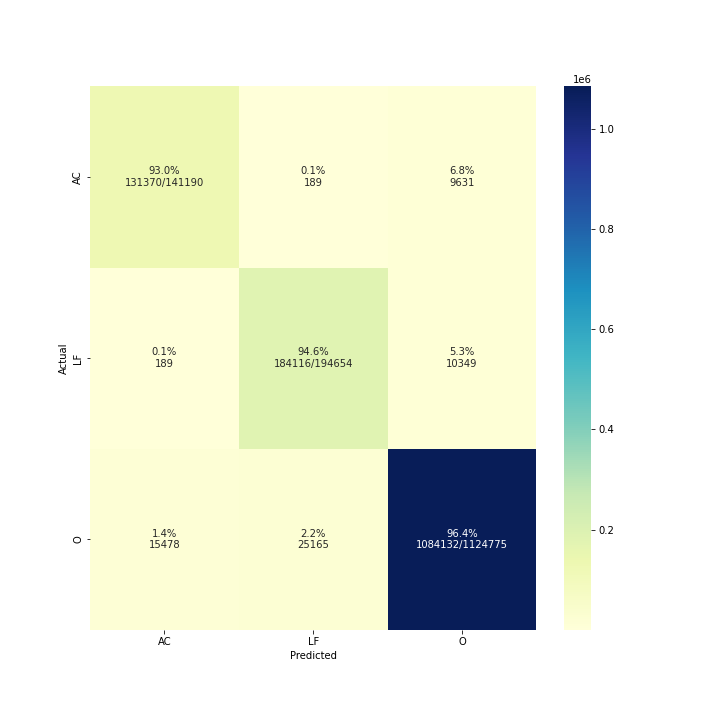}
         \caption{Confusion Matrix on PLOD\textsubscript{test-filtered}}
         \label{fig:test_filtered}
     \end{subfigure}
     \begin{subfigure}[b]{0.495\textwidth}
         \centering
         \includegraphics[width=\textwidth]{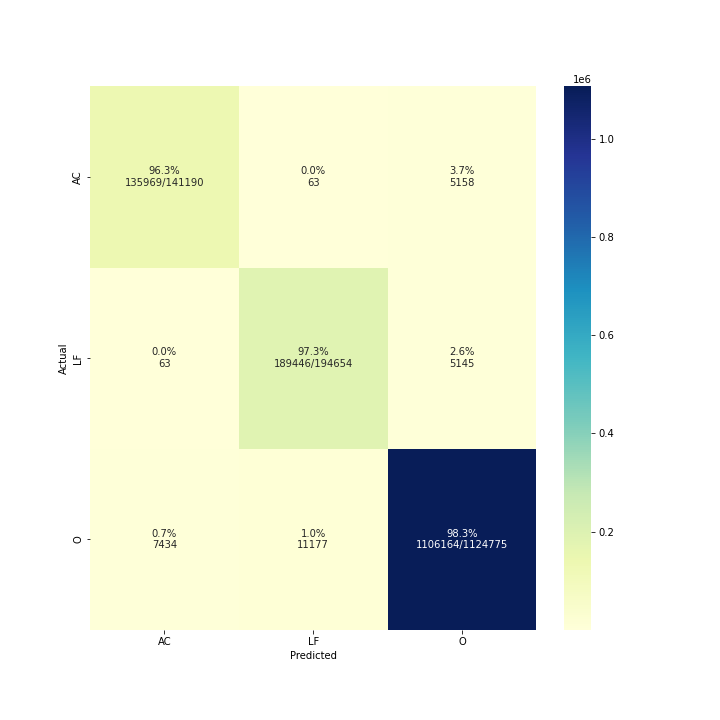}
         \caption{Confusion Matrix on PLOD\textsubscript{test-unfiltered}}
         \label{fig:test_unfiltered}
     \end{subfigure}
     \caption{Confusion Matrices over test set predictions from our best models, \textit{i.e.,} RoBERTa\textsubscript{large}}
     \label{fig:cm_abb_det}
\end{figure*}

\section{Results and Discussion}
\label{sec:results}

Based on the experiment setup discussed in Section \ref{sec:resourceeval}, we performed the evaluation of the PLOD dataset in various scenarios. Initially, we performed fine-tuning with the \textbf{PLOD\textsubscript{Unfiltered Dataset}}, and utilised nine variants of pre-trained LMs for the task. As can be seen in Table~\ref{tab:un_filtered_results}, our models are able to achieve a decent performance on both test sets, both in terms of precision and recall. We observed that the RoBERTa models seem to outperform the others, with the highest F1-scores in all of the cases. We also note that the RoBERTa\textsubscript{large} model shows significantly higher precision and recall values of 0.911 and 0.9335, with an F1-score of 0.922. 

However, when testing this model trained on our data with the train+dev set of the SDU Shared Task dataset, we see a drop in performance. We attribute this drop in performance to various reasons: (1) There were spurious annotations in the SDU dataset which had been pointed out to the task organisers earlier this year when the dataset was released; (2) the domains used in the SDU dataset are `legal' and `scientific' whereas our dataset is mostly based on the Biomedical domain; (3) the drop in performance of long forms, specifically, can be attributed to incorrect classification of some tokens in the long forms which consist of many tokens. When the models trained on the PLOD\textsubscript{Unfiltered Dataset} are tested with the \textbf{SDU shared task data}, however, the RoBERTa\textsubscript{base} model seems to detect abbreviations better (0.672) than the RoBERTa\textsubscript{large} model; but for long forms RoBERTa\textsubscript{large} model outperforms (0.264) every other model. 

In Table~\ref{tab:filtered_results}, we present the results based on the \textbf{PLOD\textsubscript{Filtered Dataset}}. It seems that the RoBERTa\textsubscript{large} model and the RoBERTa\textsubscript{base} model again perform the task with significantly higher F1-score than others. We observed a similar performance on both the datasets, and each language model is performing better individually when trained on the unfiltered datasets. Our filtration process takes out many data points from the dataset which impacts the performance of the task. Also, on both the datasets BERT\textsubscript{large-cased} models also shows a comparable performance with minor differences from our best performing model.

When the models trained on the PLOD\textsubscript{Filtered Dataset} are tested with the \textbf{SDU shared task data} (Table~\ref{tab:filtered_results}), however, it can be seen that the results are more homogeneous in terms of model performance. The RoBERTa\textsubscript{base} model shows a much better performance for both long forms and abbreviations, attaining F1-scores of 0.255 and 0.683, respectively. We also observe that these models have higher precision values compared to the other models, especially RoBERTa\textsubscript{large}, which leads to the next part of our discussion: confusion matrices. Since RoBERTa\textsubscript{large} performs well on our dataset, we show the confusion matrices obtained by fine-tuning on both Filtered and Unfiltered datasets in Figure~\ref{fig:cm_abb_det}. 

Based on Figure~\ref{fig:test_filtered}, we infer that there are a large number of abbreviations (AC tag) wrongly classified with the `O' tag (6.8\%, \textit{i.e.,} 9631 tokens), when the Filtered dataset is used. However, using the Unfiltered dataset (Figure \ref{fig:test_unfiltered}, this number is reduced to 3.7\%, \textit{i.e.,} 5158 tokens. The overall accuracy of correctly classified AC tags is 93\% and, for long forms, it is at 94.6\%. The misclassified number of long forms stands at 2.2\%, \textit{i.e.,} $\sim$ 25k tokens, which is also a large number in a real-world scenario. This clearly indicates that there is a need to improve the model performance before we apply it to a real-world Biomedical domain scenario. Again, on the Unfiltered dataset, these language models show a better performance and the misclassified long-form tokens are reduced to 1\%, \textit{i.e., } $\sim$ 11k tokens. 

From the tables and confusion matrices above, we can conclude that overall, RoBERTa models perform the best for the task of abbreviation detection. However, given the current results, we also plan to conduct further experiments which use an ensemble approach with multiple models.




\section{Conclusion and Future Work}
\label{sec:conc}

In this study, we motivated the importance of abbreviation detection as an NLP task in the scientific domain and discussed the challenges one can encounter while trying to perform this task. We collected a large number of abbreviations and their corresponding long forms from open-sourced PLOS Journals and described the data collection process in detail. With some efforts towards the validation of this data, we were able to identify problems and further filter the dataset. Based on an unfiltered and a filtered version of this dataset, we performed an extensive evaluation of the abbreviation detection task by utilising various pre-trained language models. These models were not only tested on our test data but also on the SDU@AAAI-22 acronym detection shared task dataset. By analysing the results, we showed how some state-of-the-art transformer models fare at this task. With the hopes that these models might be of importance to the NLP community, we release them publicly along with the code and the raw datasets (both filtered and unfiltered).

In the future, we plan to extend this dataset with additional sources, which can be added to our data. We also plan to extend our experiments further with an ensemble approach, which can utilise various language models to perform the detection of abbreviations and their corresponding long forms.

\section*{Acknowledgement}\label{ack}

We acknowledge the PLOS Open Access Journals and thousands of unnamed authors whose research papers and abbreviated content resulted in the creation of this dataset. 

\section*{Bibliographical References}\label{reference}

\bibliographystyle{lrec2022-bib}
\bibliography{lrec2022-example}

\end{document}